\documentclass[10pt,twocolumn,letterpaper]{article}

\usepackage{cvpr}
\usepackage{times}
\usepackage{epsfig}
\usepackage{graphicx}
\usepackage{amsmath}
\usepackage{amssymb}
\usepackage{amsthm}
\usepackage{float}
\usepackage[export]{adjustbox}
\usepackage{subcaption}
\usepackage{graphics}
\usepackage{makecell}
\usepackage{multirow}
\usepackage{flushend}
\usepackage{xcolor,soul,colortbl}

\DeclareMathAlphabet{\mathbbold}{U}{bbold}{m}{n}

\usepackage{authblk}

\usepackage[pagebackref=true,breaklinks=true,letterpaper=true,colorlinks,bookmarks=false]{hyperref}

\cvprfinalcopy

\ifcvprfinal\fi

\newtheorem{theorem}{Theorem}
\numberwithin{theorem}{section}
\newtheorem{proposition}[theorem]{Proposition}
\newtheorem{definition-proposition}[theorem]{Definition-Proposition}

\newcommand{\RR}{\mathbb{R}}

\def\comment#1{{}}

 \date{}

\newcommand{\parag}[1]{\vspace{4pt}\noindent\textbf{#1~}}

\begin{document}
\pagenumbering{gobble}

\title{Unsupervised Image Matching and Object Discovery as Optimization}
\author[1,2,3]{Huy V. Vo}
\author[1,2]{Francis Bach}
\author[4]{Minsu Cho}
\author[5]{Kai Han}
\author[6]{Yann LeCun}
\author[3]{Patrick P{\'e}rez}
\author[1,2]{Jean Ponce}

\makeatother
\renewcommand\Affilfont{\small}
\affil[1]{D{\'e}partement  d'informatique de l'ENS, ENS, CNRS, PSL University, Paris, France}
\makeatletter
\renewcommand\maketitle{\AB@maketitle} 
\renewcommand\AB@affilsepx{\quad\protect\Affilfont}

\affil[2]{INRIA, Paris, France}
\affil[3]{Valeo.ai}
\affil[4]{POSTECH}
\affil[5]{University of Oxford}
\affil[6]{New York University}

\renewcommand\Authands{ and }

\twocolumn[{
	\renewcommand\twocolumn[1][]{#1}
	\maketitle	
	\begin{center}
		\vspace{-0.5cm}
		\includegraphics[width=0.99\linewidth]{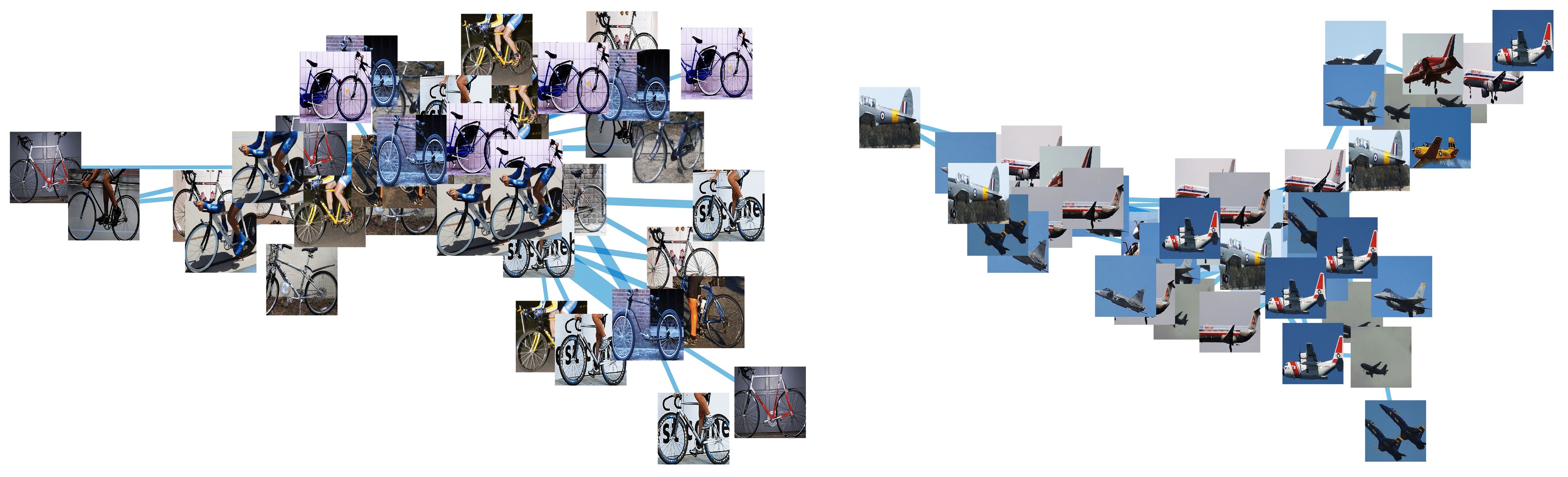}
		\captionof{figure}
		{
			{\small The proposed optimization-based method automatically discovers links between images that depict similar objects. This figure shows two image  clusters that emerge as a by-product of this approach on the VOC\_6x2 object recognition dataset that mixes 6 classes under two viewpoints. See text for details.}
		}
		\label{fig:teaser}
	\end{center}	
	\vspace{0.5cm}
}]

\begin{abstract}
Learning with complete or partial supervision is powerful but relies on ever-growing human  annotation efforts. 
As a way to mitigate this serious problem, as well as to serve specific applications, unsupervised learning has emerged
as an important field of research. In computer vision, unsupervised learning comes in various guises. We focus here on 
the unsupervised discovery and matching of object categories among images in a collection, following the work of Cho {\em et al.}~\cite{CKSP15}.
We show that the original approach can be reformulated and solved as a proper optimization problem. 

Experiments on several benchmarks establish the merit of our approach.   
\end{abstract}

\section{Introduction}
Remarkable progress has been achieved in visual
tasks such as image categorization, object detection, or semantic
segmentation, typically using fully supervised algorithms and vast
amount of manually annotated data
(e.g.,~\cite{felzenszwalb2010object,He17,He16,krizhevsky2012imagenet,Lazebnik2006,Ren15,Imagenet15}).
With the advent of crowd-sourcing, large corporations and, to a lesser
extent, academic units can launch the corresponding massive annotation
efforts for {\em specific} projects that may involve
millions images~\cite{Imagenet15}.  

But handling Internet-scale image (or video) repositories or the
continuous learning scenarios associated with personal assistants or
autonomous cars demands approaches less hungry for manual annotation.
Several alternatives are possible, including {\em weakly supervised}
approaches that rely on readily available
meta-data~\cite{Alayrac2018,Bojanowski2015} or image-level
labels~\cite{Deselaers:2010he,Joulin2010,Joulin14,Kim2011,Rubinstein2013,Tang14}
instead of more complex annotations such as bounding boxes~\cite{felzenszwalb2010object,Ren15} or
object masks~\cite{He17} as supervisory signal; {\em semi supervised}
methods~\cite{belkin2004regularization,kingma2014semi} that exploit a relatively small number of fully annotated pictures, together with a larger set of unlabelled
images; and {\em self supervised} algorithms that take advantage of
the internal regularities of image parts~\cite{DoGuEf15,NoFa16} or
video subsequences~\cite{AgCaMa15,MaCoLC16,WaGu15} to construct image
models that can be further fine-tuned in fully supervised settings. 

We address here the even more challenging problem of discovering both the
structure of image collections -- that is, which images depict similar objects (or textures, scenes, actions, etc.), and the objects
in question, in a {\em fully unsupervised}
setting~\cite{BoJo17,Caron18,faktor2012clustering,CVPR/LeeG10,Rubinstein2013,Russell06,sivic2008unsupervised}. Although
weakly, semi, and self supervised methods may provide a more {\em
  practical} foundation for large-scale visual recognition, the
fully unsupervised construction of image models is a {\em fundamental}
scientific problem in computer vision, and it should be studied. In
addition, any reasonable solution to this problem will facilitate
subsequent human labelling (by presenting discovered groups to
the operator) and scaling through automatic label
propagation, help interactive query-based visual search by linking
ahead of time fragments of potential interest, and provide a way to
learn visual models for subsequent recognition.

\subsection{The implicit structure of image collections}
Any collection of images, say, those found on the Internet, or more
modestly, in a dataset such as Pascal VOC'07, admits a natural graph
representation, where nodes are the pictures themselves, and edges
link pairs of images with similar visual content.  In {\em supervised}
image categorization
(e.g.,~\cite{krizhevsky2012imagenet,Lazebnik2006}) or object
detection (e.g.,~\cite{felzenszwalb2010object,He17,Ren15}) tasks, both the graph
structure and the visual content are clearly defined: Annotators
typically sort the images into bags, each one intended to represent
some ``object'', ``scene'' or, say, ``action'' class (``horse'',
``forest'', ``playing tennis'', etc.). Two nodes are linked by an edge
when they are associated with the same bag, and each class is
empirically defined by the images (or some manually-defined
rectangular regions within) in the corresponding connected component
of the graph. 
In {\em weakly supervised}
cosegmentation~\cite{Joulin2010,Kim2011,Rubinstein2013} or
colocalization~\cite{Deselaers:2010he,Joulin14,Tang14} tasks, on the
other hand, the graph is fully connected, and all images are supposed
to contain instances of the (few) same object categories, say,
``horse'', ``grass'', ``sky'', ``background''. Manual intervention is
reduced to selecting which images to put into a single bag, and the
visual content, in the form of regions defined by pixel-level symbolic
labels or bounding boxes associated with one of the predefined
categories, is discovered using a clustering algorithm.
\footnote{In both the cases of
  supervised image categorization/object detection and
  weakly supervised cosegmentation/colocalization, once the graph
  structure and the visual content have been identified at {\em
    training time}, these can be used to learn a model of the
  different object classes and add nodes, edges, and possibly
  additional bounding boxes at {\em test time}.}

We address in this paper the much more difficult problem of {\em fully
  unsupervised} image matching and object discovery, where both the
graph structure and a model of visual content in the form of object
bounding boxes must be extracted from the native data without {\em
  any} manual intervention. This problem has been addressed in various
forms, e.g., clustering~\cite{faktor2012clustering}\footnote{Note that plain unsupervised clustering, whether classic, spectral, discriminative or deep \cite{Diffrac,hershey2016deep,lloyd1982least,ng2002spectral}, focuses on data partitioning and not on the discovery of subsets of matching items within a cluttered collection.}, image
matching~\cite{Rubinstein2013} or topic
discovery~\cite{Russell06,sivic2008unsupervised} (see
also~\cite{BoJo17,Caron18}, where ``pseudo-object'' labels are learned
in an unsupervised manner). 
In this presentation, we build directly on
the work of Cho {\em et al.}~\cite{CKSP15} (see~\cite{Kwak2015} for
related work): Given an image and its neighbors, assumed to contain
the same object, a robust matching technique exploits both appearance
and geometric consistency constraints to assign confidence and
saliency (``stand-out'') scores to region proposals in this 
image.  The overall discovery algorithm alternates between {\em
  localization} steps where the neighbors are fixed and the regions
with top saliency scores are selected as potential objects, and {\em
  retrieval} steps where the confidence of the regions within
potential objects are used to find the nearest neighbors of each
image. After a fixed number of steps, the region with top saliency in
each image is declared to be the object it contains. Empirically, this
method has been shown in~\cite{CKSP15} to give good results. However,
it does not formulate image matching and object discovery as a proper
optimization problem, and there is no guarantee that successive
iterations will improve some objective measure of performance. The aim
of this paper is to remedy this situation.

\section{Proposed approach}
\subsection{Problem statement}

Let us consider a set of $n$ images, each containing $p_i$ rectangular
region proposals, with $i$ in $\{1\dots n\}$.  We assume that the images are
equipped with some implicit graph structure, where there is a link
between two images when the second image contains at least one object from a category depicted in the first one,
and our aim is to discover this structure, that is, find the links and the
corresponding objects.  To model this problem, let us define an
indicator variable $x_i^k$, whose value is 1 when region number $k$ of
image $i$ corresponds to a ``foreground object'' (visible in large part and from a category that occurs multiple times in the image collection), and 0 otherwise. We
collect all the variables $x_i^k$ associated with image $i$ into an
element $x_i$ of $\{0,1\}^{p_i}$, and concatenate all the variables
$x_i$ into an element $x$ of $\{0,1\}^{\sum_{i=1}^np_i}$. Likewise,
let us define an indicator variable $e_{ij}$, whose value is 1 if
image $j$ contains an object also occurring in image $i$, with $1\le
i,j\le n$ and $j\neq i$, and 0 otherwise,
collect all the variables $e_{ij}$ associated with image $i$ into an
element $e_i$ of $\{0,1\}^n$, and concatenate all the variables $e_i$
into an $n\times n$ matrix $e$ with rows $e_i^T$. Note that we can use
$e$ to define a neighborhood for each image in the set: Image $j$ is a
neighbor of the image $i$ if $e_{ij} = 1$. By definition, $e$ defines
an undirected graph if $e$ is symmetric and a directed one
otherwise. Let us also denote by $S_{ij}^{kl}$ the similarity between
regions $k$ and $l$ of images $i$ and $j$, and by $S_{ij}$ the
$p_i\times p_j$ matrix with entries $S_{ij}^{kl}$.

We propose to maximize with respect to $x$ and $e$ the objective
function
\begin{equation}
S(x,e)=\!\!\sum_{\substack{i,j=1\\j\neq i}}^n \!e_{ij}\! 
\!\sum_{\substack{1\le k\le p_i\\ 1\le l\le p_j}}\!\!\!S_{ij}^{kl} x_i^kx_j^l 
=\!\sum_{\substack{i,j=1\\j\neq i}}^n \!x_i^T [e_{ij}S_{ij}] x_j.\label{eq:main}
\end{equation}
Intuitively maximizing $S(x,e)$ encourages building edges between
images $i$ and $j$ that contain regions $k$ and $l$ with a strong
similarity $S_{ij}^{kl}$.  Of course we would like to impose certain
constraints on the $x$ and $e$ variables. The following cardinality
constraints are rather natural:\\
\noindent$\bullet$ An image should not contain more than a prededined
number of objects, say $\nu$,
\begin{equation}
\forall\,\, i\in 1\ldots n,\,\, x_i\cdot\mathbbold{1}_{p_i}\le \nu,
\label{eq:aux1}
\end{equation}
where $\mathbbold{1}_{p_i}$ is the element of $\RR^{p_i}$ with all
entries equal to one.

\noindent$\bullet$ An image should not match more than a predefined
number of other images, say $\tau$,
\begin{equation}
\forall\,\, i\in 1\ldots n,\,\,e_i\cdot\mathbbold{1}_n\le \tau.
\label{eq:aux2}
\end{equation}

\noindent{\bf Assumptions.~} We will suppose from now on that $S_{ij}$
is elementwise nonnegative, but not necessarily symmetric (the similarity model we explore in Section 3 is asymmetrical). Likewise,
we will assume that the matrix $e$ has a zero diagonal but is not
necessarily symmetric.

Under these assumptions, the cubic pseudo-Boolean function $S$ is
supermodular~\cite{BoHa02}.  Without constraints, this type of
functions can be maximized in polynomial time using a max-flow
algorithm~\cite{BiMi85} (in the case of $S(x,e)$, which does not
involve linear and quadratic terms, the solution is of course trivial
without constraints, and amounts to setting all $x_i^k$ and $e_{ij}$ with $i \neq j$ to 1).  When
the cardinality constraints (\ref{eq:aux1}-\ref{eq:aux2}) are
added, this is not the case anymore, and we have to resort to a
gradient ascent algorithm as explained next.

\subsection{Relaxing the problem}
\label{sec:optimization}
Let us first note that, for
binary variables $x_i^k$, $x_j^l$ and $e_{ij}$, we have
\begin{equation}
\label{eq:cont_main}
S(x,e)=\sum_{\substack{i,j=1\\j\neq i}}^n\sum_{\substack{1\le k\le p_i\\ 1\le l\le p_j}}
S_{ij}^{kl} \min (e_{ij}, 
x_i^k, x_j^l),
\end{equation}
with $S_{ij}^{kl}\ge 0$. Relaxing our problem so all variables are
allowed to take values in $[0,1]$, our objective becomes a sum of
concave functions, and thus is itself a concave function, defined over
the convex set (hyperrectangle) $[0,1]^N$, where $N$ is the total
number of variables. This is the standard tight
concave continuous relaxation of supermodular functions.

The Lagrangian associated with our relaxed problem is
\begin{equation}
  K(x,e;\lambda,\mu)= S(x,e)-\sum_{i=1}^n[ \lambda_i
    (x_i\cdot\mathbbold{1}_{p_i}-\nu) +\mu_i
    (e_i\cdot\mathbbold{1}_n-\tau)],
\label{eq:lag}
\end{equation}
where $\lambda=(\lambda_1,\ldots,\lambda_n)^T$ and
$\mu=(\mu_1,\ldots,\mu_n)^T$ are {\em positive} Lagrange multipliers.
The function $S(x, e)$ is concave and the primal problem is strictly
feasible; hence Slater's conditions~\cite{Slater50} hold, and we have the following
equivalent primal and dual versions of our problem
\begin{equation}
\left\{\begin{array}{l}
\max_{(x,e)\in D} \inf_{\lambda,\mu\ge 0} K(x,e;\lambda,\mu),
\\
\min_{\lambda,\mu\ge 0} \sup_{(x,e)\in D} K(x,e;\lambda,\mu),
\end{array}\right.
\label{eq:dual}
\end{equation}
where the domain $D$ is the Cartesian product of 
$[0,1]^{\sum_i p_i}$ 
and the space of
$n\times n$ matrices with entries in $[0,1]$ and a zero diagonal. With slight abuse we denote it $D=[0,1]^N$, with $N=\sum_ip_i + n(n-1)$.

\subsection{Solving the dual problem}
We propose to solve the dual problem with a subgradient descent
approach. Starting from some initial values for $\lambda^0$ and
$\mu^0$, we use the update rule
\begin{equation}
\left\{\begin{array}{l}
\lambda_i^{t+1}=[\lambda_i^t+\alpha (x_i^t\cdot\mathbbold{1}_{p_i}-\nu)]_+,\\
\mu_i^{t+1}=[\mu_i^t+\beta (e_i^t\cdot\mathbbold{1}_n-\tau)]_+,\\
\end{array}\right.
\label{eq:update_ln}
\end{equation}
where $[\cdot]_+$ denotes positive part, $k\ge 0$, $\alpha$ and $\beta$ are fixed step sizes, 
$x_i^t\cdot\mathbbold{1}_{p_i}-\nu$ and $e_i^t\cdot\mathbbold{1}_n-\tau$ are
respectively the negative of the subgradients of the Lagrangian with
respect to $\lambda_i$ and $\mu_i$ in $\lambda_i^t$ and $\mu_i^t$, and
\begin{equation}
(x^t,e^t)\in\text{argmax}_{(x,e)\in[0,1]^N} K(x,e;\lambda^t,\mu^t). 
\end{equation}

As shown in Appendix, for fixed values of $\lambda$ and $\mu$, 
our Lagrangian is a {\em supermodular} pseudo-Boolean function of 
binary variables sets $x$ and $e$.
%
%
This allows us to take
advantage of the following direct corollary of 
\cite[Prop. 3.7]{Bach13}.
\begin{proposition}
Let $f$ denote some supermodular pseudo-Boolean function of $n$
variables. We have
\begin{equation}
\max_{x\in\{0,1\}^n}f(x)=\max_{x\in[0,1]^n}f(x),
\end{equation}
and the set of maximizers of $f(x)$ in $[0,1]^n$ is the convex hull of
the set of maximizers of $f$ on $\{0,1\}^n$.
\end{proposition}
In particular, we can take
\begin{equation}
(x^t,e^t)\in\text{argmax}_{(x,e)\in\{0,1\}^N} K(x,e;\lambda^t,\mu^t).
\label{eq:max_xe}
\end{equation}
As shown in~\cite{BiMi85,BoHa02}, the corresponding
supermodular cubic pseudo-Boolean function optimization problem is
equivalent to a maximum stable set problem in a bipartite {\em
  conflict graph}, which can itself be reduced to a maximum-flow
problem. 
See Appendix for details.

Note that the size of the min-cut/max-flow problems that have to be
solved is conditioned by the number of nonzero $S_{ij}^{kl}$ entries,
which is upper-bounded by $n^2p^2$ when the matrices $S_{ij}$ are
dense (denoting $p=\max\{p_i\}$).  This is prohibitively high given that, in practice, $p$ is
between $1000$ and $4000$. To make the computations manageable, we set
all but between $100$ and $1000$ (depending on the dataset's size) of the largest entries in $S_{ij}$ to
zero in our implementation.

\subsection{Solving the primal problem}
Once the dual problem is solved, as argued by Nedi\'c \&
Ozdaglar~\cite{NeOz09} and Bach~\cite{Bach13}, an approximate solution
of the primal problem can be found as a running average of the primal
sequence $(x^t,e^t)$ generated as a by-product of the sub-gradient
method:
\vspace{-2mm}
\begin{equation}
\hat{x}=\frac{1}{T}\sum_{t=0}^{T-1} x^t,
\quad
\hat{e}=\frac{1}{T}\sum_{t=0}^{T-1} e^t
\vspace{-2mm}
\end{equation}
after some number $T$ of iterations.
Note the scalars $\hat{x}_i^k$ and $\hat{e}_{ij}$ lie in $[0,1]$ but
do not necessarily verify the constraints (\ref{eq:aux1}) and
(\ref{eq:aux2}). Theoretical guarantees on these values can be found
under additional assumptions in~\cite{Bach13,NeOz09}.

\subsection{Rounding the solution and greedy ascent}
Note that two problems remain to be solved: The solution $(\hat{x},\hat{e})$ found
now belongs to $[0,1]^N$ instead of $\{0,1\}^N$, and it may not
satisfy the original constraints. Note, however, that because of the
form of the function $S$, given some $i$ in $\{1,\ldots,n\}$ and fixed
values for $e$ and all $x_j$ with $j\neq i$, the maximum value of $S$
given the constraints is obtained by setting to 1 exactly the $\nu$
entries of $x_i$ corresponding to the $\nu$ largest entries of the
vector $\sum_{j\neq i} (e_{ij}S_{ij}+e_{ji}S_{ji}^T)x_j$. Likewise,
for some fixed value of $x$, the maximum value of $S$ is reached by
setting to 1, for all $i$ in $\{1,\ldots,n\}$, exactly the $\tau$ entries of
$e_i$ corresponding to the $\tau$ largest scalars $x_i^TS_{ij}x_j$ for
$j\neq i$ in $\{1\ldots n\}$.  This suggests the following approach to
rounding up the solution, where the variables $x_i$ are updated
sequentially in an order specified by some random permutation $\sigma$ of $\{1,\ldots,n\}$, before the variables $e_i$ are updated in parallel.
Given the permutation $\sigma$, the algorithm
below turns the running average $(\hat{x},\hat{e})$ of the primal sequence into a
discrete solution $(x,e)$ that satisfies the
conditions~(\ref{eq:aux1}) and~(\ref{eq:aux2}):
{\fbox{
\begin{minipage}{1.0\columnwidth}
\begin{tabbing}
\hspace{0.7cm}\=\hspace{0.7cm}\=\hspace{0.7cm}\=\kill
Initialize $x = \hat{x}$, $e = \hat{e}$. \\
\noindent For $i=1$ to $n$ do\\
\> Compute the indices $k_1$ to $k_\nu$ of the $\nu$ largest\\
\> elements of the vector \\ 
\>\> $\sum_{j \neq \sigma(i)}^n (e_{\sigma(i)j}S_{\sigma(i)j} + e_{j\sigma(i)}S_{j\sigma(i)}^T)x_j$.\\
\> $x_{\sigma(i)}\leftarrow 0$.\\
\> For $t=1$ to $\nu$ do $x_{\sigma(i)}^{k_t}\leftarrow 1$.\\
For $i=1$ to $n$ do\\
\> Compute the indices $j_1$ to $j_\tau$ of the $\tau$ largest scalars\\
\> $x_{i}^T S_{ij} x_{j}$.\\
\> $e_i\leftarrow 0$.\\
\> For $t=1$ to $\tau$ do $e_{ij_t}\leftarrow 1$.\\
\noindent Return $x, e$.
\end{tabbing}
\end{minipage}}}\\
	
Note that there
is no preferred order for the image indices. This actually suggests
repeating this procedure with different random permutations until the
variables $x$ and $e$ do not change anymore or some limit on the
number of iterations is reached. This iterative procedure can be seen
as a greedy ascent procedure over the discrete variables of
interest. Note that by construction the terms in the left and right
sides of (\ref{eq:aux1}) and (\ref{eq:aux2}) are equal at the optimum.

\subsection{Ensemble post processing}
\label{subsec:ensemble_method}

The parameter $\nu$ can be seen from two different 
viewpoints: (1) as the maximum number of objects that may be depicted in 
an image, or (2) as an upper bound on the total number of object region 
{\em candidates} that are under consideration in a picture. Both 
viewpoints are equally valid but, following Cho \etal~\cite{CKSP15}, we 
focus in the rest of this presentation on the second one, and present in 
this section a simple heuristic for selecting one final object region 
among these candidates. Concretely, since using random permutations 
during greedy ascent provides a different solution for each run of our 
method, we propose to apply an {\em ensemble method} to stabilize the 
results and boost performance in this selection process, itself viewed 
as a post-processing stage separate from the optimization part.

Let us suppose that after $L$ independent executions of the greedy ascent
step, we obtain $L$ solutions $(x(l), e(l)),\ 1 \leq l \leq L$. We start by combining these solutions into a
single discrete pair $(\bar{x}, \bar{e})$ where $\bar{x}$ and
$\bar{e}$ satisfy
\begin{itemize}
    \vspace{-1mm}
	\item $\bar{x}_i^k = 1$ if $\exists \ l, 1 \leq l \leq L$ such that $x_i^k(l) = 1$,
	\vspace{-1mm}
	\item $\bar{e}_{ij} = 1$ if $\exists \ l, 1 \leq l \leq L$ such that $e_{ij}(l) = 1$. 
	\vspace{-1mm}
\end{itemize}
This way of combining the individual solutions can be seen as a \textit{max pooling} procedure. We have also tried average pooling but found it less effective. Note that after this intermediate step, an image might violate any of the two constraints (\ref{eq:aux1}-\ref{eq:aux2}). This is not a problem in this postprocessing stage of our method. Indeed, we next show how to use $\bar{x}$ and $\bar{e}$ to select a {\em single} object proposal for each image.


We choose a single proposal for each image out of those retained in $\bar{x}$ (proposals $(i,k)$ s.t. $\bar{x}_i^k=1$). To this end, we rank the proposals in image $i$ according to a score $u_i^k$ defined for each proposal $(i,k)$ as
\vspace{-1mm}
\begin{equation}
u_i^k = \bar{x}_i^k \sum_{j \in \mathcal{N}(i,k)} \max_{l | \bar{x}_j^l=1} S_{ij}^{kl},
\label{eq:sim}
\vspace{-1mm}
\end{equation}
where 
$\mathcal{N}(i,k)$ is composed of the $\tau$ images represented by the $1s$ in $\bar{e}_i$ which have the largest similarity to $(i,k)$ as measured by $\max_{l | \bar{x}_j^l=1} S_{ij}^{kl}$. Finally, we choose the proposal in image $i$ with maximum score $u_i^k$ as the final object region. Note that the graph of images corresponding to these final object regions can be retrieved by computing $e$ that maximizes the objective function given the value of $x$ defined by these regions as in the greedy ascent. Also, the method above can be generalized to more than one proposal per image using the defined ranking.


\section{Similarity model\label{sec:similarity_model}}

Let us now get back to the definition of the similarity function
$S_{ij}$. As advocated by Cho {\em et al.}~\cite{CKSP15}, a
rectangular region which is a tight fit for a compact object (the {\em
  foreground}) should better model this object than a larger region,
since it contains less {\em background}, or than a smaller region (a
{\em part}) since it contains more foreground. Cho {\em et al.}~\cite{CKSP15} only
implement the first constraint, in the form of a {\em stand-out}
score. We discuss in this section how to implement these ideas in the
optimization context of this work.

\subsection{Similarity score}
Following \cite{CKSP15}, the similarity score
between proposal $k$ of image $i$ and proposal $l$ of image
$j$ can be defined as
\begin{equation}
\label{eq:similarity_score_PHM}
s_{ij}^{kl} = 
a_{ij}^{kl} \sum_{o\in O} 
g(r_i^k,r_j^l,o) 
\!\!\!\sum_{\substack{1 \leq k' \leq p_i \\ 1 \leq l' \leq p_j}} \!\!\!g(r_i^{k'},r_j^{l'},o) a_{ij}^{k'l'},
\vspace{-2mm}
\end{equation}
where $a_{ij}^{kl}$ is a similarity term based on appearance alone,
using the WHO descriptor (whiten HOG)~\cite{dalal2005histograms, hariharan2012who} in our case,
$r_i^k$ and $r_j^l$ denote the image rectangles associated with the
two proposals, $o$ is a discretized offset (translation plus two scale
factors) taking values in $O$, and $g(r,s,o)$ measures the geometric
compatibility between $o$ and the rectangles $r$ and $s$.
Intuitively, $s_{ij}^{kl}$ scales the appearance-only score
$a_{ij}^{kl}$ by a geometric-consistency term akin to a generalized
Hough transform~\cite{Ballard81}, see~\cite{CKSP15} for details.

Note that we can rewrite Eq.~(\ref{eq:similarity_score_PHM}) as
\begin{equation}
s_{ij}^{kl}=b_{ij}^{kl}\cdot c_{ij},
\end{equation}
where $b_{ij}^{kl}$ is the vector of dimension $|O|$ with entries
$a_{ij}^{kl}g(r_i^k,r_j^l,o)$, and $c_{ij}=\sum_{k',l'=1}^p
b_{ij}^{k'l'}$.  
The $p_ip_j$ vectors $b_{ij}^{kl}$ and the vector $c_{ij}$ can be
precomputed with time and storage cost of ${\cal O}(p^2|O|)$. Each
term $s_{ij}^{kl}$ can then be computed in ${\cal O}(|O|)$ time, and
the matrix $S_{ij}$ can thus be computed with a total time and space
complexity of ${\cal O}(p^2|O|)$.

Note that the score $s_{ij}^{kl}$ defined by
Eq.~(\ref{eq:similarity_score_PHM}) depends on the number of
region proposals per images, which may introduce a bias for
edges between images that contain many region proposals. It may
thus be desirable to {\em normalize} this score by defining it instead as
\begin{equation}
\label{eq:similarity_score_PHM_normalized}
s_{ij}^{kl}=\dfrac{1}{p_ip_j}b_{ij}^{kl}\cdot c_{ij}.
\vspace{-2mm}
\end{equation}

\subsection{Stand-out score}
Let us identify the region proposals contained in some image $i$ with
their index $k$, and define $P_i^k$ as the set of regions that are
{\em parts} of that region (that is, they are included, with some
tolerance, within $k$). Let us also define $B_i^k$ as the set of
regions that form the {\em background} for $k$ (that is, $k$ is
included, with some tolerance, within these regions). 
Let $r_i^k$ denote the actual rectangular image region associated with
proposal $k$ in image $i$, and let $A(r)$ denote the area of some
rectangle $r$. A plausible definition for $P_i^k$ is
\begin{equation}
P_i^k=\{l\,:\,A(r_i^k\cap r_i^l)> \rho A(r_i^l) \},
\end{equation}
for some reasonable value of $\rho$, e.g., 0.5.
Likewise, a plausible definition for $B_i^k$ is
\begin{equation}
B_i^k =\{l: A(r_i^k\cap r_i^l)> \delta A(r_i^k) \,\,\text{and}
\,\, A(r_i^l)> \gamma A(r_i^k)\},
\end{equation}
for reasonable values of $\delta$ and $\gamma$, e.g., 0.8 and 2.
Following~\cite{CKSP15}, we define the {\em stand-out score} of a match
$(k,l)$ as
\begin{equation}
\label{eq:standout_score}
S_{ij}^{kl}=s_{ij}^{kl}-v_{ij}^{kl},\,\,\text{where}\,\,
v_{ij}^{kl}=\max_{(k',l')\in B_i^k\times B_j^l}
s_{ij}^{k'l'}.
\end{equation}
With this definition, $S_{ij}^{kl}$ may be negative. In our implementation,
we threshold 
these scores so they are nonnegative.

When $B_i^k$ and $B_j^l$ are large, which is generally the case when
the regions $r_i^k$ and $r_j^l$ are small, a brute-force computation
of $v_{ij}^{kl}$ may be very slow. We propose below instead a simple
heuristic that greatly speeds up calculations.

Let $Q_{ij}$ denote the set formed by the $q$ matches $(k,l)$ with
highest scores $s_{ij}^{kl}$, sorted in increasing order, which can be
computed in ${\cal O}(p^2\log p)$.  The stand-out scores can be
computed efficiently by the following procedure:
\fbox{
\begin{minipage}{1.0\columnwidth}
\begin{tabbing}
\hspace{0.7cm}\=\hspace{0.7cm}\=\hspace{0.7cm}\=\kill
Initialize all $v_{ij}^{kl}$ to $0$.\\
For each match $(k',l')$ in $Q_{ij}$ do\\
\> For each match $(k,l)$ in $P_i^{k'}\times P_j^{l'}$ do
$v_{ij}^{kl}=s_{ij}^{k'l'}$.\\
For $k=1$ to $p_i$ and $l=1$ to $p_j$ do\\
\> If $s_{ij}^{kl}>0$ and $v_{ij}^{kl}=0$ then 
$v_{ij}^{kl}=\displaystyle\!\!\max_{(k',l')\in B_i^k\times B_j^l}\!\!s_{ij}^{k'l'}$.
\end{tabbing}
\end{minipage}
}\\

The idea is  that relatively few high-confidence matches $(k',l')$ in $Q_{ij}$
can be used to efficiently compute many stand-out scores. There is
a trade-off between the cost of this step, ${\cal O}(\sum_{(k',l')\in Q_{ij}}
|P_i^{k'}|\,|P_j^{l'}|)$, and the number of variables $v_{ij}^{kl}$ it assigns
a value to, ${\cal O}(|\cup_{(k',l') \in Q_{ij}}P_i^{k'}\times P_j^{l'}|)$. In practice, we have
found that taking $q=10,000$ is a good compromise, with only about 5\%
of the stand-out scores being computed in a brute-force manner, and
a significant speed-up factor of over 10.

\section{Experiments and results}
\paragraph{Datasets, proposals and metric.} 

For our experiments we use the same datasets ({\em ObjectDiscovery} [OD],
{\em VOC\_6x2} and {\em VOC\_all}) and region proposals (obtained by
the {\em randomized Prim's algorithm} [RP]~\cite{manen2013prime}) as Cho {\em et al.}~\cite{CKSP15}.  
OD consists of pictures of three object classes (\textit{airplane},
\textit{horse} and \textit{car}) with outliers not containing any
object instance. There are 100 images per category, with 18, 7 and 11
outliers respectively (containing no object instance).
VOC\_all is a subset of the PASCAL VOC2007
train$+$val dataset obtained by eliminating all images containing only
objects marked as \textit{difficult} or \textit{truncated}. Finally,
VOC\_6x2 is a subset of VOC\_all containing only images of 6
classes -- {\em aeroplane}, {\em bicycle}, {\em boat}, {\em bus}, {\em
  horse} -- and {\em motorbike} from two different views, \textit{left}
and \textit{right}.

For evaluation, we use the standard \textit{CorLoc} measure, the percentage of images
correctly localized. It is a proxy metric in the case of unsupervised discovery.
An image is ``correctly localized'' when the intersection over union ($IoU$) between one
of the ground-truth regions and the predicted one is greater than
0.5. Following~\cite{CKSP15}, we evaluate our algorithm in ``separate'' 
and ``mixed'' settings. In the former case, the class-wise performance is averaged over classes. In the latter, a single performance is computed over all classes jointly. In our experiments, we use $\nu=5$, $\tau=10$ and standout matrices with 1000 non-zero entries unless mentioned otherwise.

\parag{Separate setting.}
We firstly evaluate different settings of our
algorithm on the two smaller datasets, OD and VOC\_6x2. The
performance is governed by three design choices: (1) using the normalized stand-out score ({\em NS)} or its unnormalized version, (2) using continuous optimization ({\em CO}) or variables $x$ and $e$ with all entries equal to one to initialize the greedy ascent procedure, and (3) using the ensemble method ({\em EM}) or not. In total, we thus have eight configurations to test.

\begin{table}
\centering
\resizebox{0.35\textwidth}{!}{%
\begin{tabular}{|c|c|c|c|c|}
\hline
\multicolumn{3}{|c|}{Method} & OD & VOC\_6x2 \\ \hline \hline
\multicolumn{3}{|c|}{\makecell{Cho {\em et al.}}} 
& 84.2 & 67.7 \\ \hline
\multicolumn{3}{|c|}{\makecell{Cho {\em et al.}, our version}} 
& 84.2 & 67.6 \\ \hline
\hline
\multirow{4}{*}{w/o EM} & 
\multirow{2}{*}{w/o CO} 
& w/o NS & 81.9 $\pm$ 0.9 & 65.9 $\pm$ 1.0 \\ \cline{3-5}
&  & w NS & 83.1 $\pm$ 0.8 & 67.2 $\pm$ 1.0 \\ \cline{2-5}
 & \multirow{2}{*}{w CO} & w/o NS & 82.9 $\pm$ 
0.8 & 66.6 $\pm$ 0.7 \\ \cline{3-5}
 &  & w NS & 84.4 $\pm$ 0.8 & 68.1 $\pm$ 0.9 \\ \hline
\hline
\multirow{4}{*}{w EM} & 
\multirow{2}{*}{w/o CO} & w/o NS & 84.4 $\pm$ 
0.0 & 68.8 $\pm$ 0.4 \\ \cline{3-5}
 &  & w NS & 85.6 $\pm$ 0.3 & 
68.7 $\pm$ 0.5 \\ \cline{2-5}
 & \multirow{2}{*}{w CO} & w/o NS & 83.8 
$\pm$ 0.2 & 67.4 $\pm$ 0.4 \\ \cline{3-5}
 &  & w NS & \textbf{85.8 $\pm$ 0.6} & \textbf{69.4 $\pm$ 0.3} \\ \hline
\end{tabular}
}
\vspace{-2mm}
\caption{\small Performance of different configurations of our
  algorithm compared to the results of 
  Cho {\em et al.} on Object Discovery 
  and VOC\_6x2 datasets in the separate setting.}
\label{table:performance_od_voc_6x2_separate}
\vspace{-4mm}
\end{table}

The results are shown in
Table~\ref{table:performance_od_voc_6x2_separate}. We have found a small bug in the publicly available code
of Cho \etal~\cite{CKSP15}, and report both the results from~\cite{CKSP15} and those we
obtained after correction. We observe that the normalized
standout score always gives comparable or better results than its
unnormalized counterpart, while the ensemble method also improves both the
score and the stability (lower variance) of our solution. Combining the normalized standout score, the ensemble method, and the continuous optimization initialization to greedy ascent yields the best performance. 
Our best results outperform \cite{CKSP15} by small but statistically significant margins: 1.6\% for OD and  1.8\% for VOC\_6x2.
Finally, to assess the merit of the continuous optimization, we have measured its duality gap on OD and VOC\_6x2: it ranges from 1.5\% to 8.7\% of the energy, with an average of 5.2\% and 3.9\% on  the two datasets respectively.

\begin{table}
\centering
\resizebox{0.25\textwidth}{!}{%
\begin{tabular}{|c|c|c|}
\hline
\multicolumn{2}{|c|}{Method} & VOC\_all \\ \hline \hline
\multicolumn{2}{|c|}{Cho {\em et al.}} & 36.6 \\ \hline
\multicolumn{2}{|c|}{Cho {\em et al.}, our execution} & 37.6 \\ \hline \hline
\multirow{2}{*}{w/o CO} & w/o EM & 36.4 $\pm$ 0.3 \\ \cline{2-3}
 & w EM & 39.0 $\pm$ 0.2 \\ \hline \hline 
\multirow{2}{*}{w CO} & w/o EM & 37.8 $\pm$ 0.3 \\ \cline{2-3}
 & w EM & 39.2 $\pm$ 0.2 \\ \hline \hline 
\multicolumn{2}{|c|}{Li {\em et al.} ~\cite{Li2016}} & 40.0 \\ \hline
\multicolumn{2}{|c|}{Wei {\em et al.} ~\cite{Wei2017}} & \textbf{46.9} \\ \hline
\end{tabular}
}
\vspace{-2mm}
\caption{\small Performance on VOC\_all in separate setting with different configurations.
}
\label{table:performance_voc_all_separate}
\vspace{-6mm}
\end{table}

We now evaluate our algorithm on VOC\_all. As the complexity of solving the max flow problem grows very fast with the number of images, for configurations with continuous optimization, we reduce the number of non-zero entries in each standout matrix such that the total number of nodes in the graph is around $2 \times 10^7$. These standout matrices are then used in rounding the continuous solution, but in the greedy ascent procedure we switch to standout matrices with 1000 non-zero entries. For configurations without the continuous optimization, we always use the standout matrices with 1000 non-zero entries. Also, to reduce the memory footprint of our method, we prefilter the set of potential neighbors of each image for the class {\em person} that contains 1023 pictures. Pre-filtering is done by marking 100 nearest neighbors of each image in terms of Euclidean distance between GIST \cite{torralba2008small} descriptors as potential neighbors. In the separate setting, we only apply the pre-filtering on the class \textit{person} which has 1023 images. The other classes are sufficiently small for not resorting to the prefiltering procedure.

Table \ref{table:performance_voc_all_separate} shows the CorLoc values obtained by our method with different configurations compared to Cho \etal. It can be seen that the ensemble postprocessing and the continuous optimization are also helpful on this dataset. We obtain the best result with the configuration that includes both of them, which is 1.6\% better than Cho \etal. However, our performance is still inferior to state of the art in image colocalization \cite{Li2016, Wei2017} which employ deep features from convolutional neural networks trained for image classification and explicitly exploits the single-class assumption. 

\parag{Mixed setting.} 
We now compare in Table \ref{table:performance_mixed} the performance of our algorithm to Cho \etal in the mixed setting (none of the other methods is applicable to this case). It can be seen that our algorithm without the continuous optimization has the best performance among those in consideration. Compared to Cho \etal, it gives a CorLoc 0.8\% better on OD dataset, 4.3\% better on VOC\_6x2 and 2.3\% better on VOC\_all. The decrease in performance of our method when using the continuous optimization is likely due to the fact that we use standout matrices with only 200 non-zero entries on OD, 100 non-zero entries on VOC\_6x2 and 100 non-zero entries on VOC\_all (due to the limit on the number of nodes of the bipartite graphs) in the configuration with the continuous optimization while we use standout matrices with 1000 non-zero entries in the configuration without the continuous optimization.  

\begin{table}
\centering
\resizebox{0.4\textwidth}{!}{%
\begin{tabular}{|c|c|c|c|}
\hline
Method & OD & VOC\_6x2 & VOC\_all \\ \hline \hline

Cho {\em et al.} & - & - & 37.6 \\ \hline
Cho {\em et al.}, our execution & 82.2 & 55.9 & 37.5 \\ \hline \hline

w/o CO & \textbf{83.0 $\pm$ 0.4} & \textbf{60.2 $\pm$ 0.4} & \textbf{39.8 $\pm$ 0.2} \\ \hline
w CO  & 80.8 $\pm$ 0.5 & 59.3 $\pm$ 0.4 & 38.5 $\pm$ 0.2 \\ \hline
\end{tabular}
}
\vspace{-2mm}
\caption{\small Performance on the datasets in mixed setting.}
\label{table:performance_mixed}
\end{table}

\parag{Sensitivity to $\nu$.} 
We compare the performance of our method when using different values of $\nu$ on the VOC\_6x2 dataset.\footnote{Note that we have also tried the interpretation of $\nu$ as the maximum number of objects per image, without satisfying results so far.} Table \ref{table:performance_voc_6x2_separate_nu_1_5} shows the CorLoc obtained by different configurations of our algorithm, all with normalized standout. The performance consistently increases with the value of $\nu$ on this dataset. In all other experiments however, we set $\nu=5$ to ease comparisons to~\cite{CKSP15}.
\begin{table}
\centering
\resizebox{0.25\textwidth}{!}{%
\begin{tabular}{|c|c|c|c|}
\hline
\multicolumn{3}{|c|}{Method} & VOC\_6x2 \\ \hline \hline

\multirow{4}{*}{$\nu=1$} & \multirow{2}{*}{w/o CO } & w/o EM & 63.5 $\pm$ 1.2 \\ \cline{3-4}
 &  & w EM & 67.7 $\pm$ 0.8 \\ \cline{2-4}
 & \multirow{2}{*}{w CO} & w/o EM & 65.8 $\pm$ 0.8 \\ \cline{3-4}
 &  & w EM & \textbf{68.1 $\pm$ 0.7} \\ \hline \hline

\multirow{4}{*}{$\nu=5$} & \multirow{2}{*}{w/o CO } & w/o EM & 67.2 $\pm$ 1.0 \\ \cline{3-4}
 &  & w EM & 68.7 $\pm$ 0.5 \\ \cline{2-4}
 & \multirow{2}{*}{w CO} & w/o EM & 68.1 $\pm$ 0.9 \\ \cline{3-4}
 &  & w EM & \textbf{69.4 $\pm$ 0.3} \\ \hline \hline 
 
\multirow{4}{*}{$\nu=10$} & \multirow{2}{*}{w/o CO } & w/o EM & 68.6 $\pm$ 1.0 \\ \cline{3-4}
 &  & w EM & 69.1 $\pm$ 0.3 \\ \cline{2-4}
 & \multirow{2}{*}{w CO} & w/o EM & 68.9 $\pm$ 0.7 \\ \cline{3-4}
 &  & w EM & \textbf{70.0 $\pm$ 0.3} \\ \hline
 
\end{tabular}
}
\vspace{-2mm}
\caption{\small Performance of different configurations of our algorithm with $\nu=1$, $\nu=5$ and $\nu=10$.}
\label{table:performance_voc_6x2_separate_nu_1_5}
\vspace{-4mm}
\end{table}

\parag{Using deep features.}
Since activations from deep neural networks trained for image classification (deep features) are known to be better image representations than handcrafted features in various tasks, we have also experimented with such descriptors. We have replaced WHO~\cite{hariharan2012who} by activations from different layers in VGG16 \cite{Simonyan14c}, when computing the appearance similarity between regions. In this case, the similarity between two regions is simply the scalar product of the corresponding deep features (normalized or not). As a preliminary experiment to evaluate the effectiveness of deep features, we have run our algorithm without the continuous optimization with the standout score computed using layers \textit{conv4\_3}, \textit{conv5\_3} and \textit{fc6} in VGG16. 
Table \ref{table:performance_voc_6x2_deep_separate} shows the results of these experiments. Surprisingly, most of the deep features tested give worse results than WHO. This may be due to the fact that our matching task is more akin to image retrieval than classification, for which deep features are typically trained. Among those tested, only a variant of the features extracted from the layer \textit{conv5\_3} of VGG16 gives an improvement (about 2\%) compared to the result obtained by using WHO.
\begin{table}[H]
\centering
\resizebox{0.35\textwidth}{!}{%
\begin{tabular}{|c|c|c|c|}
\hline
\multicolumn{3}{|c|}{Features} & Average \\ \hline \hline

\multicolumn{3}{|c|}{WHO} & 68.8 $\pm$ 0.5 \\ \hline \hline

\multirow{4}{*}{\textit{conv4\_3}} & \multirow{2}{*}{\makecell{warping + \\ center cropping}} & unnormalized & 64.2 $\pm$ 0.2 \\ \cline{3-4} 
 & & normalized & 57.1 $\pm$ 0.6 \\ \cline{2-4}
 & \multirow{2}{*}{ROI pooling~\cite{girshickICCV15fastrcnn}} &  unnormalized & 63.1 $\pm$ 0.2 \\ \cline{3-4} 
 & & normalized & 63.4 $\pm$ 0.4 \\ \hline
 
\multirow{4}{*}{\textit{conv5\_3}} & \multirow{2}{*}{\makecell{warping + \\ center cropping}} & unnormalized & 64.9 $\pm$ 0.2 \\ \cline{3-4} 
 & & normalized & 64.1 $\pm$ 0.4 \\ \cline{2-4}
 & \multirow{2}{*}{ROI pooling~\cite{girshickICCV15fastrcnn}} & unnormalized & \textbf{70.7 $\pm$ 0.2} \\ \cline{3-4}
 & & normalized & 68.2 $\pm$ 0.3 \\ \hline

\multirow{2}{*}{\textit{fc6}} &\multirow{2}{*}{\makecell{warping + \\ center cropping}} & unnormalized & 61.3 $\pm$ 0.2 \\ \cline{3-4} 
 & & normalized & 61.0 $\pm$ 0.4 \\ \hline
 
\end{tabular}
}
\vspace{-2mm}
\caption{\small Performance of our algorithm with deep features on VOC\_6x2 in the separate setting.}
\label{table:performance_voc_6x2_deep_separate}
\vspace{-4mm}
\end{table}

\paragraph{Unsupervised initial proposals.} It should be noted that, although our algorithm like that of Cho \etal~\cite{CKSP15}  is totally unsupervised once {\em given the region
  proposals}, the randomized Prim's algorithm itself is
supervised~\cite{manen2013prime}. To study the effect of this built-in supervision, we have also
tested the unsupervised {\em selective search}
algorithm~\cite{UijlingsIJCV2013} for choosing region proposals. We
have conducted experiments on VOC\_6x2 dataset with the three
different settings of selective search ({\em fast}, {\em medium} and
{\em quality}). As one might expect, the {\em fast} mode gives
the smallest number of proposals and of \textit{positive} ones (proposals
whose $IoU$ with one ground truth box is greater than 0.5); the {\em
  quality} mode outputs the largest set of proposals and of positive ones, 
  the {\em medium} mode lies in-between. To compare with~\cite{CKSP15}, we also run their public software with each mode of
selective search.

\begin{table}[H]
\centering
\resizebox{0.35\textwidth}{!}{%
\begin{tabular}{|c|c|c|c|}
\hline
\multicolumn{2}{|c|}{Proposal algorithm} & Cho {\em et al.} & Ours \\
\hline
\multirow{3}{*}{selective search} & \textit{fast} & 23.3 & 41.4 $\pm$ 0.5 \\ \cline{2-4}
 & \textit{medium} & 20.6 & 48.4 $\pm$ 0.5 \\ \cline{2-4} 
 & \textit{quality} & 32.6 & 62.8 $\pm$ 0.6 \\
\hline
\multicolumn{2}{|c|}{randomized Prim's} & 67.6 & 69.4 $\pm$ 0.4 \\
\hline
\end{tabular}}
\vspace{-2mm}
\caption{\small Object discovery on VOC\_6x2 with selective search and
  randomized Prim's as region proposal algorithms.}
\label{table:eval_vocx_ss}
\vspace{-3mm}
\end{table}

The results are shown in
Table~\ref{table:eval_vocx_ss}. It can be seen that the performance of both Cho \etal's method and ours drop significantly when using selective search. This may be due to the fact that the percentage of positive proposals found by selective search is much smaller than that of RP. However, we see that with the {\em quality} mode of selective
search, our method gives results quite close to those of 
RP, whereas the method in~\cite{CKSP15} fails badly.
This suggests that our method is more robust.

\parag{Visualization.} 
In order to gain insight into the structures discovered by our approach, we derive from its output a graph of image regions and visualize its main connected components. The nodes of this graph are the image regions that have been finally retained. Two regions $(i,k)$ and $(j,l)$ are connected if the images containing them are neighbors in the discovered undirected image graph ($e_{ij}$ or $e_{ji} = 1$) and the standout score between them, $S_{ij}^{kl}$, is greater than a certain threshold.
   
Choosing the threshold to get a sufficient number of large enough components for visualization purpose has proven difficult. We used instead an iterative procedure: the graph is first constructed with a high threshold to produce a small number of connected components of reasonable size, which are removed from the graph. On the remaining graph, a new, suitable threshold is found to get new components of sufficient size. This is repeated until a target number of components is reached. 

\begin{figure}
\begin{center}
\includegraphics[width=\linewidth]{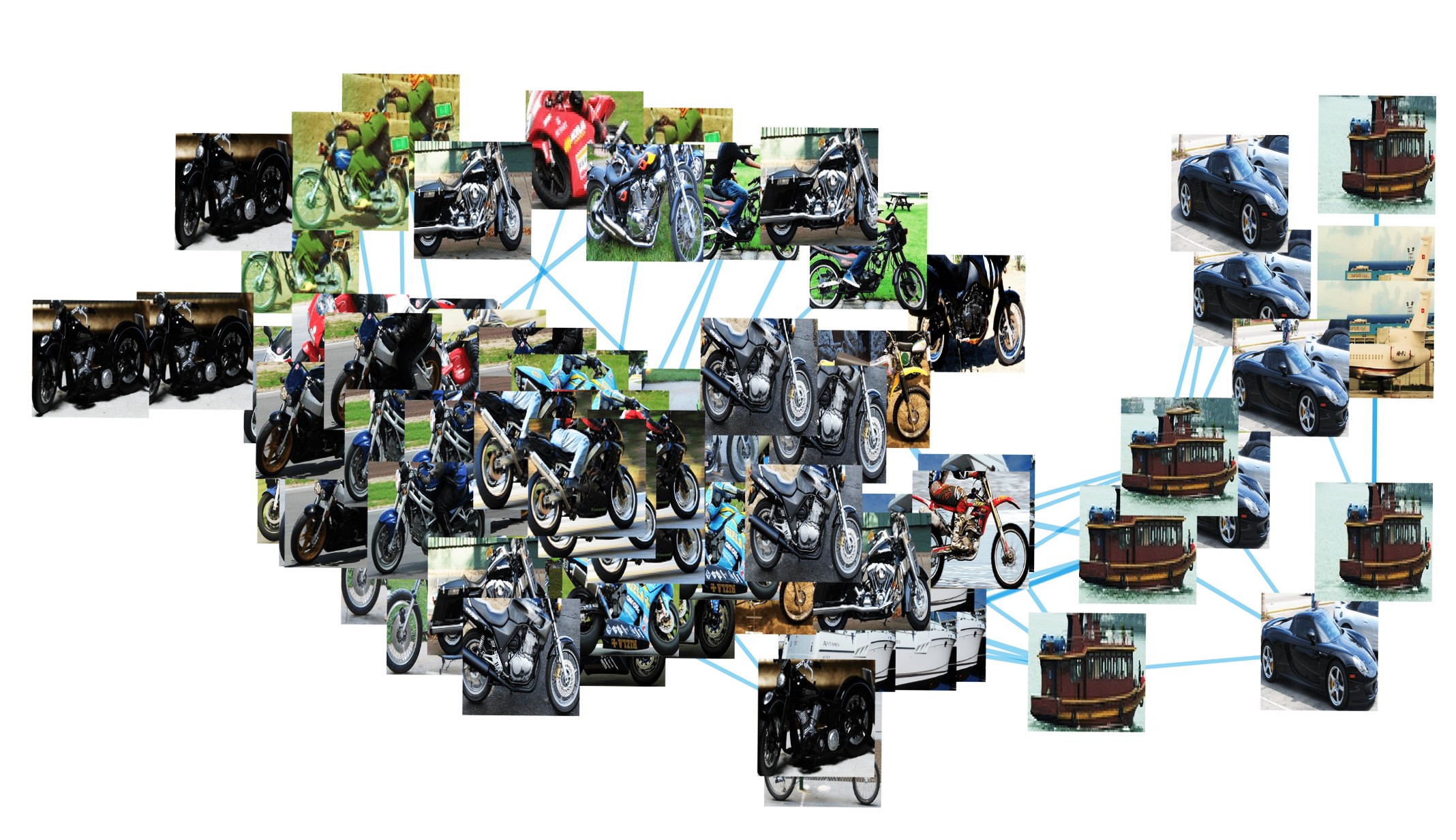}
\end{center}
\vspace{-6mm}
\caption{\small Visualization of VOC\_6x2 in the mixed setting. The figure shows the third component in the graph of regions, corresponding roughly to class \textit{motorbike}. The two first components are shown in Fig.\ref{fig:teaser}.}
\label{fig:component3}
\vspace{-4mm}
\end{figure}

When applied to our results in the mixed setting on VOC\_6x2 dataset, this visualization procedure yields clusters that roughly match object categories. In Figure \ref{fig:teaser}, we show sub-sampled graphs (for visualization purpose) of the two first components, which roughly correspond to classes \textit{bicycle} and \textit{aeroplane}. The third component is shown in Figure \ref{fig:component3}. Although containing also images of other classes, it is by far dominated by \textit{motorbike} images. The visualization suggests that our model does extract meaningful semantic structures from the image collections and regions they contain.

\section{Conclusion}
We have presented an optimization-based approach to fully unsupervised image
matching and object discovery and demonstrated its promise on several standard benchmarks. In its current form, our algorithm is limited to relatively small datasets. We are exploring several paths for scaling up its performance, including better mechanisms based on deep features and the PHM algorithm for pre-filtering image neighbors and selecting regions proposals. Future work will also be dedicated to developing effective ensemble methods for discovering multiple objects in images, further investigating a symmetric version of the proposed approach using an undirected graph, understanding why deep features do not give better results in our context, and improving our continuous optimization approach so as to handle large datasets in a mixed setting, perhaps through some form of variable clustering.
\vspace{-2mm}
\section*{Appendix: Maximization of supermodular cubic pseudo-Boolean functions} 
An immediate corollary of \cite[Lemma 1]{BiMi85} is
that a cubic pseudo-Boolean function with nonegative trinary coefficients
and no binary terms is supermodular.  For fixed 
$\lambda$
and $\mu$, this is obviously the case for the Lagrangian $K$ in (\ref{eq:lag}).

In addition, the unary terms in $K$ are nonpositive, and the Langragian
can thus be rewritten, up to some constant additive term, in the form
\begin{equation}
f(x_1,\ldots,x_n)=
\sum_{i\in U} c_i\bar{x}_i
+\displaystyle\sum_{(i,j,k)\in T} c_{ijk}x_ix_jx_k,
\label{eq:cubicps}
\end{equation}
where $\bar{x}_i=1-x_i$ (the {\em complement} of $x_i$),
$U\subset\{1,\ldots,n\}$, $T\subset\{1,\ldots,n\}^2$, and all
coefficients $c_i$ and $c_{ijk}$ are positive.  We specialize in the
rest of this section the general maximization method of~\cite{BiMi85}
to functions of this form.

The {\em conflict graph}~\cite{BiMi85,BoHa02} $G(f)$ associated with
such a function $f$ has as a set of nodes $X(f)=V\cup W$, where the
elements of $V$ correspond to linear terms, those of $W$ correspond to
cubic terms, and an edge links to nodes when one of the corresponding
terms contains a variable, and the other one its complement. By
construction $G(f)$ is a bipartite graph, with edges joining only
elements of $V$ to elements of $W$.

As shown in~\cite{BiMi85} maximizing $f$ amounts to finding a maximum
weight stable set in $G(f)$, where the nodes of $V$ are assigned
weights $c_i$ and the nodes of $W$ are assigned weights $c_{ijk}$,
which in turn reduces to computing a maximum flow between nodes $s$
and $t$ in the network deducted from $G(f)$ by (1) adding a source
node and edges with upper capacity bound $c_i$ between $s$ and the
corresponding elements of $V$; (2) adding a sink node $t$ and edges
with upper capacity bound $c_{ijk}$ between the corresponding elements
of $W$ and $t$; (3) assigning to all edges (from $V$ to $W$) in $G(f)$
an upper capacity bound of $+\infty$.

Let $[A,\bar{A}]$ denote the minimum cut obtained by computing the
maximum flow in this graph, where $s$ is an element of $A$ and $t$ is
an element of $\bar{A}=X(f)\setminus A$. The maximum weight stable set
is then $S=(A\cap V)\cup (\bar{A}\cap W)$. The monomials $\bar{x}_i$
and $x_ix_jx_k$ associated with elements of $S$ are set to 1, from
which the values of all variables are easily deduced.
\vspace{-2mm}
\vspace{-2mm}
\paragraph{Acknowledgments.} This work was supported in part by the Inria/NYU collaboration agreement, the Louis Vuitton/ENS chair on artificial intellgence and the EPSRC Programme Grant Seebibyte EP/M013774/1. We also thank Simon Lacoste-Julien for his valuable comments and suggestions.

\clearpage

\begin{thebibliography}{10}\itemsep=-1pt

\bibitem{AgCaMa15}
P.~Agrawal, J.~Carreira, and J.~Malik.
\newblock Learning to see by moving.
\newblock In {\em ICCV}, 2015.

\bibitem{Alayrac2018}
J.-B. Alayrac, P.~Bojanowski, N.~Agrawal, I.~Laptev, J.~Sivic, and
  S.~Lacoste-Julien.
\newblock Learning from narrated instruction videos.
\newblock {\em IEEE Trans. Pattern Anal. and Machine Intell.},
  40(9):2194--2208, 2018.

\bibitem{Bach13}
F.~Bach.
\newblock Learning with submodular functions: A convex optimization
  perspective.
\newblock {\em Foundations and Trends in Machine Learning}, 6(2-3):145--373,
  2013.

\bibitem{Diffrac}
F.~Bach and Z.~Harchaoui.
\newblock {DIFFRAC} : a discriminative and flexible framework for clustering.
\newblock In {\em Proc. Neural Info. Proc. Systems}, 2007.

\bibitem{Ballard81}
D.~Ballard.
\newblock Generalizing the {H}ough transform to detect arbitrary shapes.
\newblock {\em Pattern Recognition}, 1981.

\bibitem{belkin2004regularization}
M.~Belkin, I.~Matveeva, and P.~Niyogi.
\newblock Regularization and semi-supervised learning on large graphs.
\newblock In {\em COLT}, 2004.

\bibitem{BiMi85}
A.~Billionnet and M.~Minoux.
\newblock Maximizing a supermodular pseudoboolean function: A polynomial
  algorithm for supermodular cubic functions.
\newblock {\em Discrete Applied Mathematics}, 12:1--11, 1985.

\bibitem{BoJo17}
P.~Bojanowski and A.~Joulin.
\newblock Unsupervised learning by predicting noise.
\newblock In {\em ICML}, 2017.

\bibitem{Bojanowski2015}
P.~Bojanowski, R.~Lajugie, E.~Grave, F.~Bach, I.~Laptev, J.~Ponce, and
  C.~Schmid.
\newblock Weakly-supervised alignment of video with text.
\newblock In {\em ICCV}, 2015.

\bibitem{BoHa02}
E.~Boros and P.~Hammer.
\newblock Pseudo-{B}oolean optimization.
\newblock {\em Discrete Applied Mathematics}, 123(1-3):155--225, 2002.

\bibitem{Caron18}
M.~Caron, P.~Bojanowski, A.~Joulin, and M.~Douze.
\newblock Deep clustering for unsupervised learning of visual features.
\newblock In {\em ECCV}, 2018.

\bibitem{CKSP15}
M.~Cho, S.~Kwak, C.~Schmid, and J.~Ponce.
\newblock Unsupervised object discovery and localization in the wild:
  Part-based matching with bottom-up region proposals.
\newblock In {\em CVPR}, 2015.

\bibitem{dalal2005histograms}
N.~Dalal and B.~Triggs.
\newblock Histograms of oriented gradients for human detection.
\newblock In {\em CVPR}, 2005.

\bibitem{Deselaers:2010he}
T.~Deselaers, B.~Alexe, and V.~Ferrari.
\newblock Localizing objects while learning their appearance.
\newblock In {\em ECCV}, 2010.

\bibitem{DoGuEf15}
C.~Doersch, A.~Gupta, and A.~Efros.
\newblock Unsupervised visual representation learning by context prediction.
\newblock In {\em ICCV}, 2015.

\bibitem{faktor2012clustering}
A.~Faktor and M.~Irani.
\newblock Clustering by composition--unsupervised discovery of image
  categories.
\newblock In {\em ECCV}, 2012.

\bibitem{felzenszwalb2010object}
P.~Felzenszwalb, R.~Girshick, D.~McAllester, and D.~Ramanan.
\newblock Object detection with discriminatively trained part-based models.
\newblock {\em IEEE Trans. Pattern Anal. and Machine Intell.},
  32(9):1627--1645, 2010.

\bibitem{girshickICCV15fastrcnn}
R.~Girshick.
\newblock Fast {R}-{CNN}.
\newblock In {\em ICCV}, 2015.

\bibitem{hariharan2012who}
B.~Hariharan, J.~Malik, and D.~Ramanan.
\newblock Discriminative decorrelation for clustering and classification.
\newblock In {\em ECCV}, 2012.

\bibitem{He17}
K.~He, G.~Gkioxari, P.~Dollar, and R.~Girshick.
\newblock Mask {R-CNN}.
\newblock In {\em ICCV}, 2017.

\bibitem{He16}
K.~He, X.~Zhang, S.~Ren, and J.~Sun.
\newblock Deep residual learning for image recognition.
\newblock In {\em CVPR}, 2016.

\bibitem{hershey2016deep}
J.~R. Hershey, Z.~Chen, J.~Le~Roux, and S.~Watanabe.
\newblock Deep clustering: Discriminative embeddings for segmentation and
  separation.
\newblock In {\em ICASSP}, 2016.

\bibitem{Joulin2010}
A.~Joulin, F.~Bach, and J.~Ponce.
\newblock {Discriminative clustering for image co-segmentation}.
\newblock In {\em CVPR}, 2010.

\bibitem{Joulin14}
A.~Joulin, K.~Tang, and L.~Fei-Fei.
\newblock Efficient image and video co-localization with {Frank-Wolfe}
  algorithm.
\newblock In {\em ECCV}, 2014.

\bibitem{Kim2011}
G.~Kim and E.~Xing.
\newblock {Distributed cosegmentation via submodular optimization on
  anisotropic diffusion}.
\newblock In {\em ICCV}, 2011.

\bibitem{kingma2014semi}
D.~P. Kingma, S.~Mohamed, D.~J. Rezende, and M.~Welling.
\newblock Semi-supervised learning with deep generative models.
\newblock In {\em Proc. Neural Info. Proc. Systems}, 2014.

\bibitem{krizhevsky2012imagenet}
A.~Krizhevsky, I.~Sutskever, and G.~E. Hinton.
\newblock Imagenet classification with deep convolutional neural networks.
\newblock In {\em NIPS}, 2012.

\bibitem{Kwak2015}
S.~Kwak, M.~Cho, I.~Laptev, J.~Ponce, and C.~Schmid.
\newblock Unsupervised object discovery and tracking in video collections.
\newblock In {\em ICCV}, 2015.

\bibitem{Lazebnik2006}
S.~Lazebnik, C.~Schmid, and J.~Ponce.
\newblock {Beyond bags of features: spatial pyramid matching for recognizing
  natural scene categories}.
\newblock In {\em CVPR}, 2006.

\bibitem{CVPR/LeeG10}
Y.~J. Lee and K.~Grauman.
\newblock Object-graphs for context-aware category discovery.
\newblock In {\em CVPR}, 2010.

\bibitem{Li2016}
Y.~Li, L.~Liu, C.~Shen, and A.~Hengel.
\newblock Image co-localization by mimicking a good detector's confidence score
  distribution.
\newblock In {\em ECCV}, 2016.

\bibitem{lloyd1982least}
S.~Lloyd.
\newblock Least squares quantization in {PCM}.
\newblock {\em IEEE Trans. on information theory}, 28(2):129--137, 1982.

\bibitem{manen2013prime}
S.~Manen, M.~Guillaumin, and L.~Van~Gool.
\newblock Prime object proposals with randomized {P}rim's algorithm.
\newblock In {\em ICCV}, 2013.

\bibitem{MaCoLC16}
M.~Matthieu, C.~Couprie, and Y.~LeCun.
\newblock Deep multi-scale video prediction beyond mean square error.
\newblock In {\em ICLR}, 2016.

\bibitem{NeOz09}
A.~Nedi\'c and A.~Ozdaglar.
\newblock Approximate primal solutions and rate analysis for dual subgradient
  methods.
\newblock {\em SIAM Journal on Optimization}, 19(4), 2009.

\bibitem{ng2002spectral}
A.~Y. Ng, M.~I. Jordan, and Y.~Weiss.
\newblock On spectral clustering: Analysis and an algorithm.
\newblock In {\em NIPS}, 2002.

\bibitem{NoFa16}
M.~Noroozi and P.~Favaro.
\newblock Unsupervised learning of visual representations by solving jigsaw
  puzzles.
\newblock In {\em ECCV}, 2106.

\bibitem{Ren15}
S.~Ren, K.~He, R.~Girshick, and J.~Sun.
\newblock Faster {R-CNN}: Towards real-time object detection with region
  proposal networks.
\newblock In {\em NIPS}, 2015.

\bibitem{Rubinstein2013}
M.~Rubinstein and A.~Joulin.
\newblock {Unsupervised Joint Object Discovery and Segmentation in Internet
  Images}.
\newblock In {\em CVPR}, 2013.

\bibitem{Imagenet15}
O.~Russakovsky, J.~Deng, H.~Su, J.~Krause, S.~Satheesh, S.~Ma, Z.~Huang,
  A.~Karpathy, A.~Khosla, M.~Bernstein, A.~Berg, and L.~Fei-Fei.
\newblock {ImageNet} large scale visual recognition challenge.
\newblock {\em Int. J. Computer Vision}, 115(3):211--252, 2015.

\bibitem{Russell06}
B.~Russell, W.~Freeman, A.~Efros, J.~Sivic, and A.~Zisserman.
\newblock Using multiple segmentations to discover objects and their extent in
  image collections.
\newblock In {\em CVPR}, 2006.

\bibitem{Simonyan14c}
K.~Simonyan and A.~Zisserman.
\newblock Very deep convolutional networks for large-scale image recognition.
\newblock {\em CoRR}, abs/1409.1556, 2014.

\bibitem{sivic2008unsupervised}
J.~Sivic, B.~C. Russell, A.~Zisserman, W.~T. Freeman, and A.~A. Efros.
\newblock Unsupervised discovery of visual object class hierarchies.
\newblock In {\em CVPR}, 2008.

\bibitem{Slater50}
M.~Slater.
\newblock Lagrange multipliers revisited.
\newblock {\em Cowles Commission Discussion Paper No. 403}, 1950.

\bibitem{Tang14}
K.~Tang, A.~Joulin, and L.-j. Li.
\newblock Co-localization in real-world images.
\newblock In {\em CVPR}, 2014.

\bibitem{torralba2008small}
A.~Torralba, R.~Fergus, and Y.~Weiss.
\newblock Small codes and large image databases for recognition.
\newblock In {\em CVPR}, 2008.

\bibitem{UijlingsIJCV2013}
J.~R.~R. Uijlings, K.~E.~A. van~de Sande, T.~Gevers, and A.~W.~M. Smeulders.
\newblock Selective search for object recognition.
\newblock {\em IJCV}, 2013.

\bibitem{WaGu15}
X.~Wang and A.~Gupta.
\newblock Unsupervised learning of visual representations using videos.
\newblock In {\em ICCV}, 2015.

\bibitem{Wei2017}
X.~Wei, C.~Zhang, Y.~Li, C.~Xie, J.~Wu, C.~Shen, and Z.~Zhou.
\newblock Deep descriptor transforming for image co-localization.
\newblock In {\em IJCAI}, 2017.

\end{thebibliography}
 \ifx\URL\undefined \def\URLset#1{{\tt #1}\catcode`\~=\active\catcode`\_=8}
  \def\URL{\catcode`\~=12 \catcode`\_=12 \URLset} \fi

\end{document}